\documentclass[letterpaper]{article} 
\usepackage{aaai23}  
\usepackage{times}  
\usepackage{helvet}  
\usepackage{courier}  
\usepackage[hyphens]{url}  
\usepackage{graphicx} 
\usepackage{natbib}  
\usepackage{caption} 
\usepackage[utf8]{inputenc} 
\usepackage{hyperref}       
\usepackage{booktabs}       
\usepackage{amsfonts}       
\usepackage{nicefrac}       
\usepackage{microtype}      
\usepackage{fancyhdr}       
\usepackage{tabularx}
\graphicspath{{figures/}}     
\usepackage{subfig}
\usepackage{caption}
\usepackage{multirow}

\usepackage[utf8]{inputenc}
\usepackage{authblk}
\usepackage{amsmath,amssymb,bm}
\usepackage[normalem]{ulem}
\usepackage[dvipsnames]{xcolor}
\usepackage{bbm}
\usepackage{diagbox}

\usepackage[ruled,vlined]{algorithm2e}

\newcommand{\dataDist}{\mathcal{D}}
\newcommand{\reals}{\mathbb{R}}

\newcommand{\groundTruth}{\bm{y}}

\newcommand{\Loss}[4]{\mathcal{L}_\text{#4}(#1, #2; #3)}

\newcommand{\imageSpace}{\mathcal{X}}

\newcommand{\clip}[2]{\underset{#1}{\text{clip}}\left( #2 \right)}
\newcommand{\quotes}[1]{``#1"}

\title{Robust Transferable Feature Extractors:\\Learning to Defend Pre-Trained Networks Against White Box Adversaries}
\author{Alex Cann, Ian Colbert, Ihab Amer\\\textit{Advanced Micro Devices, Inc.}}
\makeatother

\nocopyright

\begin{document}

\maketitle

\begin{abstract}

The widespread adoption of deep neural networks in computer vision applications has brought forth a significant interest in adversarial robustness. Existing research has shown that maliciously perturbed inputs specifically tailored for a given model (\textit{i.e.}, adversarial examples) can be successfully transferred to another independently trained model to induce prediction errors. 
Moreover, this property of adversarial examples has been attributed to features derived from predictive patterns in the data distribution.
Thus, we are motivated to investigate the following question: \textit{Can adversarial defenses, like adversarial examples, be successfully transferred to other independently trained models?}
To this end, we propose a deep learning-based pre-processing mechanism, which we refer to as a robust transferable feature extractor (RTFE).
After examining theoretical motivation and implications, we experimentally show that our method can provide adversarial robustness to multiple independently pre-trained classifiers that are otherwise ineffective against an adaptive white box adversary. Furthermore, we show that RTFEs can even provide one-shot adversarial robustness to models independently trained on different datasets.

\end{abstract}

\section{Introduction}
\label{sec:introduction}

As deep neural networks have become increasingly common in security-conscious applications (\textit{e.g.}, autonomous driving or malware detection), protecting deployed networks from the manipulations of a malicious attacker has become increasingly important.
It has been shown that neural networks trained to classify images using standard learning objectives are vulnerable to imperceptible pixel-level perturbations crafted by an intelligent adversary.
The algorithms used to approximate maximally destructive perturbations are commonly referred to as \textit{adversarial attacks} and are studied under formally defined \textit{threat models} consisting of sets of assumed constraints on the adversary. While various attacks have been extensively studied under a variety of threat models, we focus exclusively on defending against a \textit{white box adversary}, where the attacker has full knowledge of the model under attack and any defenses applied, as this is the most difficult scenario to defend against~\cite{carlini2019evaluating}.

Existing state-of-the-art defenses against white box adversaries induce robustness by training on adversarial examples generated throughout the learning process, a method often referred to as \quotes{adversarial training}~\cite{madry2017towards}.
While the body of work surrounding adversarial training is rapidly expanding, there are several common drawbacks:
(1) adversarial training is far more computationally expensive than the standard training paradigm~\cite{madry2017towards};
(2) adversarially trained models are known to be significantly less accurate on unattacked images than their non-robust counterparts~\cite{su2018robustness, tsipras2018robustness, raghunathan2019adversarial}; and
(3) when compared to non-robust models, adversarially trained models are more dependent on model capacity to achieve improved performance, further increasing their cost~\cite{madry2017towards}.
In this work, we explore the use of a deep neural network adversarially trained to be a reusable pre-processing defense against a white box adversary for the purpose of amortizing the costs of adversarial training over multiple non-robust models that are otherwise undefended.

\begin{figure*}[ht]
    \centering
    \includegraphics[width=0.85\linewidth]{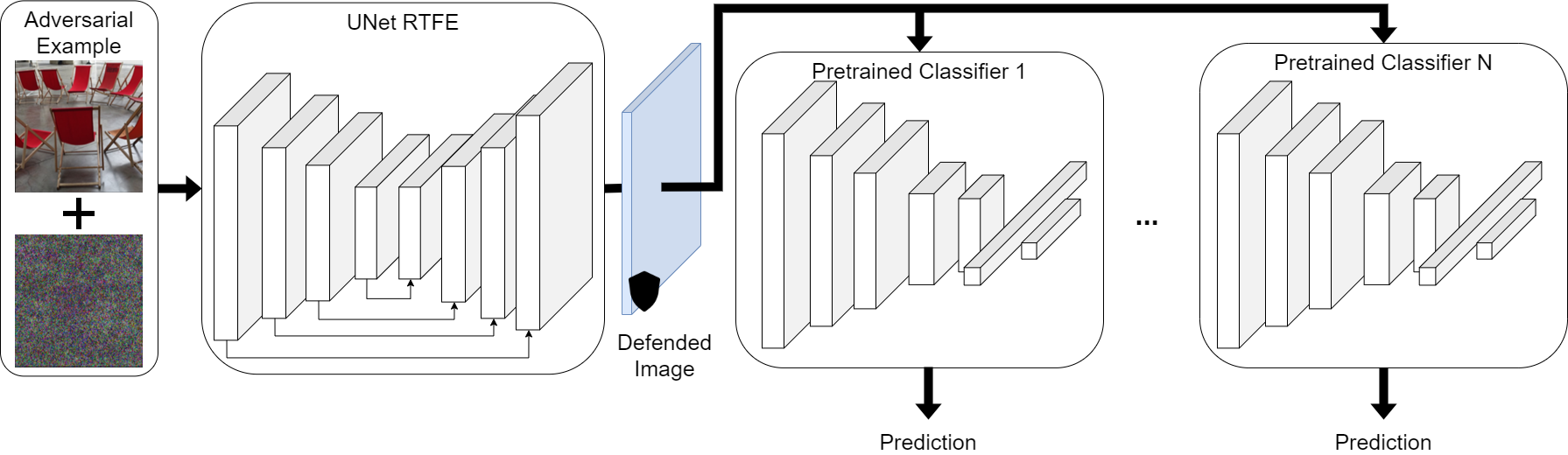}
    \caption{Our robust transferable feature extractor (RTFE) takes as input an adversarial example and gives as output a defended image that can be passed to independently pre-trained classifiers that are otherwise undefended.
    This reusable pre-processing defense is effective against an adaptive white box adversary even across neural architectures and datasets.}
    \label{fig:use_rtfe}
\end{figure*}

We are motivated in part by previous studies that have shown that adversarial examples exploit two types of vulnerabilities: \textit{features} and \textit{bugs}~\cite{ilyas2019adversarial,nakkiran2019discussion}.
While bugs represent a unique vulnerability in an individual model, such as a poor decision boundary, features represent meaningful and predictive elements of a dataset that can be exploited under perturbation.
Furthermore, the existence of feature vulnerabilities has motivated the transferability of adversarial examples~\cite{ilyas2019adversarial}.
Thus, we hypothesize that this transferability property extends to defenses designed to correct for such vulnerabilities.
More formally, we hypothesize that one can train a neural network to take as input an adversarially perturbed image and give as output a robust transferable representation with meaningful and predictive features that are useful across independently pre-trained and undefended classifiers.
We refer to this network as a robust transferable feature extractor (RTFE), visualized in Figure~\ref{fig:use_rtfe}.

While recent work has begun to investigate the usage of neural networks trained as pre-processing defenses~\cite{liao2018defense, zhou2021improving}, the resulting models either: (1) are ineffective against a white box adversary, as shown in~\cite{athalye2018obfuscated}; or (2) overfit to the model they are trained to defend, making them unable to successfully transfer, as shown in Section~\ref{sec:experiments}.
In this work, we build from the theory of~\cite{ilyas2019adversarial} and design a method to train an RTFE to defend independently pre-trained classifiers against a white box adversary.
Furthermore, we investigate the transferability of the resulting RTFEs across both models and datasets.
Finally, we build from the theory of~\cite{nakkiran2019discussion} by using model ensembling to more reliably exploit feature vulnerabilities for the purpose of increasing the transferability of RTFEs.
To the best of our knowledge, we are the first to provide a transferable pre-processing defense that is effective against a white box adversary.

\section{Background and Related Work}
\label{sec:related_work}

One of the most widely accepted defense methods against white box adversaries is the adversarial training approach of~\cite{madry2017towards}, of which there are numerous variations \cite{croce2021robustbench}.
Adversarial training replaces the standard learning objective with the robust optimization objective given by Eq.~\ref{eq:robustness_prior}, which encourages a model, parameterized by $\theta$, to minimize a loss function on all datapoints $(\bm{x}, \bm{y})$ in the training set $\dataDist$, while also being locally stable within a bounded neighborhood around each point.
\begin{equation}
\min_{\theta} ~\mathbb{E}_{(\bm{x}, \bm{y}) \sim \dataDist} \left[ \max_{\Vert \delta \Vert_p \leq \epsilon} \Loss{\bm{x} + \delta}{\groundTruth}{\theta}{}\right]
\label{eq:robustness_prior}
\end{equation}

\newcommand{\xiter}[1]{\hat{\bm{x}}_{#1}}
\newcommand{\PGDLoss}{\nabla_{\xiter{i}} \mathcal{L}_\text{CE} \left(\xiter{i},y;\theta\right)} 
\newcommand{\sign}[1]{\text{sign}\left( #1 \right)}

Noticeably, this min-max formulation is computationally intractible; thus, the problem is often decoupled in practice.
Standard approaches use projected gradient descent (PGD) to estimate the maximally worst-case perturbation $\delta$ of image $\bm{x}$ under a constraint of the form $\Vert \delta \Vert_p \leq \epsilon$, where $\Vert \cdot \Vert_p$ denotes the $\ell_p$ vector norm. The learned weights $\theta$ are then updated to minimize the standard learning objective on the perturbed images~\cite{madry2017towards}.
While PGD generalizes across any choice of bounded norm, we primarily focus our attention on adversaries bounded by the $\ell_\infty$-norm.
Our adversarial examples are first initialized from a random uniform distribution, as given by Eq.~\ref{eq:pgd_init}. We then use Eq.~\ref{eq:pgd} to iteratively generate adversarial examples, where $\alpha$ is the step size, and $\nabla_{\xiter{i}} \mathcal{L}$ is the local gradient with respect to the input image $\xiter{i}$ of the cross entropy loss $\mathcal{L}_\text{CE}$. 
\begin{align}
    \xiter{0} &= \bm{x} + \mathcal{U}(-\epsilon, \epsilon) \label{eq:pgd_init}\\
	\xiter{i+1} &= \clip{\left[\bm{x} - \epsilon, \bm{x} + \epsilon\right]}{\xiter{i}+ \alpha \cdot \sign{\PGDLoss}} \label{eq:pgd}
\end{align}

As an alternative to adversarial training methods, there have been many attempts to design pre-processing functions to defend against adversarial examples~\cite{guo2018countering, liao2018defense, zhou2021improving, joshi2022defense}.
\cite{guo2018countering} investigated a variety of traditional image pre-processing techniques (\textit{e.g.}, JPEG compression) for the purpose of removing malicious perturbations from adversarial examples; however, these techniques were shown to rely on a phenomenon referred to as \textit{gradient obfuscation} that was later bypassed by~\cite{athalye2018obfuscated}, demonstrating that these types of defenses are ineffective against a white box adversary.
\cite{liao2018defense} train an autoencoder on a static dataset of adversarial examples extracted from an undefended pre-trained classifier.
By viewing an adversarial example as an image corrupted by carefully tailored noise, they are motivated to train an autoencoder as an image denoiser for the purpose of removing adversarial perturbations.
However,~\cite{Athalye2018OnTR} show that this procedure also fails to provide a defense against an adversarial attack directed at the composition of the model and denoiser. \cite{zhou2021improving} train a denoiser to defend against a white box adversary by adversarially training a pre-processing autoencoder jointly with their classifier. We find that their method has two key limitations: (1) it requires a pre-trained robust classifier, which precludes the need for such a pre-processing defense; and (2) it does not provide robustness when re-used to defend a pre-trained classifier that was not adversarially trained, as further shown in Section~\ref{sec:experiments}.

Finally, independent of our research,~\cite{joshi2022defense} study the impact of pre-processing defenses in the domain of automatic speech recognition.
Among their set of experiments, they report favorable results for their pre-processing defense trained to provide adversarial robustness against a white box adversary for an otherwise undefended model; however, they do not explore the transferability of preprocessing defenses, which is the main focus of this work.

\section{Robust Transferable Feature Extractors}
\label{sec:rtfe}

Here, we detail our theoretical motivation and defense mechanism.
In the following sections, we denote a natural image as $\bm{x}$, its ground truth class as $\bm{y}$, any adversarial perturbation as $\delta$, and an adversarial example as $\hat{\bm{x}}= \bm{x}+\delta$.
We represent a given neural network model with learnable parameters $\theta$ as $f_\theta$ and an ensemble of $N$ models as $\{f_{\theta_i} \}_{i=1}^N$, where $i \in \{ 1, \cdots, N \}$.
Finally, We denote a robust transferable feature extractor (RTFE) with learnable parameters $\phi$ as $g_\phi$.

\subsection{Motivating Transferable Defenses}
\label{sec:motivation}

\cite{ilyas2019adversarial} theorize that the success of adversarial examples can be attributed to two types of vulnerabilities: (1) \quotes{features}, which are characteristics of images that represent useful, but unstable, directions for learning; and (2) \quotes{bugs}, which are vulnerabilities or aberrations in the model corresponding to irrelevant (and sometimes detrimental) directions for learning. Formally, they define a feature as a quantifiable quality of an image, represented mathematically as any function $f :\imageSpace \rightarrow \reals$ mapping the space of images $\imageSpace$ to the space of real numbers $\reals$.
These features can be further sub-divided into two classes: vulnerable and robust.
Whereas the correlation between \textit{vulnerable} features and the true class of an image can be inverted under perturbation, the correlation between \textit{robust} features and the true class of an image will remain the same~\cite{ilyas2019adversarial}.

Whereas \quotes{bugs} represent unique vulnerabilities of an individual neural network, \quotes{features} are meaningful and predictive elements of the data distribution.
Thus, the exploitation of vulnerable features motivates the success of transferable adversarial examples, as different models independently trained on the same dataset are likely to rely on similar sets of features to discriminate between images~\cite{ilyas2019adversarial, nakkiran2019discussion}. 
Therefore, we hypothesize that this property also motivates the  transferability of a pre-processing defense trained to either remove vulnerable features from an image or to extract robust ones.
Furthermore, given that a pre-processing defense can provide protection against feature vulnerabilities for one model, it may also defend against these same vulnerabilities present in other models independently trained on similar datasets.

Consider a natural image $\bm{x}$ composed of both robust features ($\bm{x}_\text{robust}$) and vulnerable features ($\bm{x}_\text{vulnerable}$) such that $\bm{x}=\mathcal{I}(\bm{x}_\text{vulnerable},\bm{x}_\text{robust})$, where $\mathcal{I}(\cdot)$ is some composite function that creates an image.
Our goal is to train RTFE $g_\phi$ to take image $\bm{x}$ as input and give as output a transferable representation free of vulnerable features such that $g_\phi(\bm{x}) \approx \bm{x}_\text{robust}$.
Given that $\bm{x}_\text{robust}$ contains meaningful and predictive elements that delineate the data distribution, $g_\phi(\bm{x})$ should be able to provide adversarial robustness while maintaining predictive performance for independently pre-trained models that are otherwise undefended.

\subsection{Learning a Transferable Defense Mechanism}
\label{sec:method}

\begin{figure}
\centering
\includegraphics[width=0.8\columnwidth]{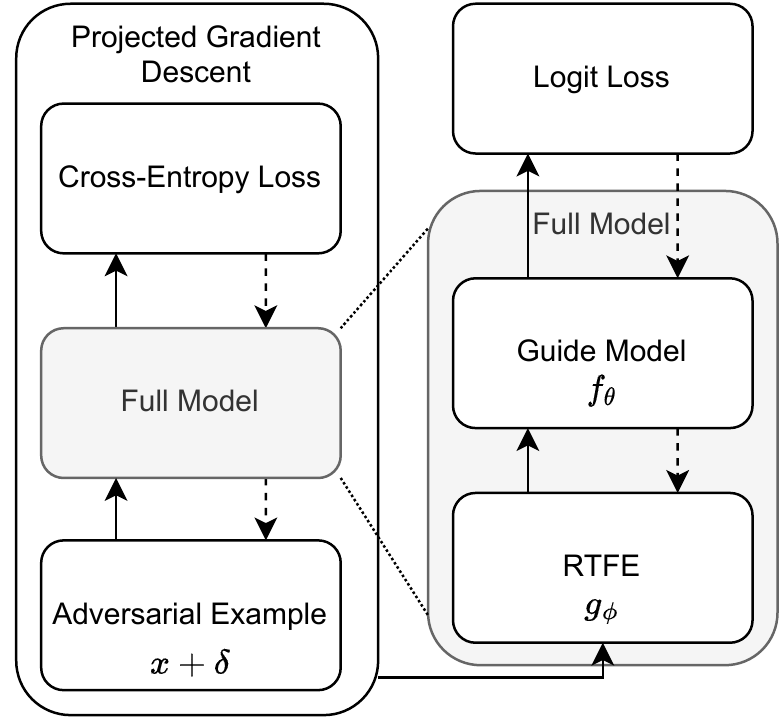}
\caption{The high-level procedure to train an RTFE $g_\phi$ using a single guide model $f_\theta$ on PGD adversarial examples.}
\label{fig:train_rtfe}
\end{figure}

To train an RTFE $g_\phi$, we build from the standard adversarial training procedure proposed by~\cite{madry2017towards}.
However, rather than training a robust classifier to learn $\theta$ by minimizing the optimization objective given by Eq.~\ref{eq:robustness_prior}, we adapt their method to learn $\phi$ by minimizing the optimization objective given by Eq.~\ref{eq:our_learning_objective} using the logit loss given by Eq.~\ref{eq:logit_loss}\footnote{We found the logit loss to be more performant when compared to the standard cross entropy loss proposed in~\cite{madry2017towards}. We also tested several auxiliary losses, such as the variations of the pixel loss from~\cite{liao2018defense}, but found that they did not impact performance. We omit them from our method for simplicity.}.
We employ the Huber loss \cite{huber1992robust}, given in Eq.~\ref{eq:huber_loss}, as our logit loss due to the increased stability in the presence of extreme outliers, which are natural artifacts of adversarial training. In our experiments, we use $\beta=1$. To generate logit values during training, we use a frozen classifier $f_\theta$, which we refer to as a \textit{guide model}, that is pre-trained using the standard learning objective. 
\begin{equation}
\min_{\phi} ~\mathbb{E}_{(\bm{x}, \bm{y}) \sim \dataDist} \left[ \max_{\Vert \bm{x} - \hat{\bm{x}} \Vert_\infty \leq \epsilon} \Loss{\hat{\bm{x}}}{\bm{x}}{\phi, \theta}{logit}\right]
\label{eq:our_learning_objective}
\end{equation}
\begin{equation}
\Loss{\hat{\bm{x}}}{\bm{x}}{\phi, \theta}{logit} = \text{huber}_\beta \left(f_\theta\left(g_\phi(\hat{\bm{x}})\right), f_\theta(\bm{x})\right)
\label{eq:logit_loss}
\end{equation}
\begin{equation}
\text{huber}_\beta (\bm{a}, \bm{b}) = \begin{cases}
\frac{1}{2}\Vert \bm{a} - \bm{b} \Vert_2^2  \hfill \text{if } \Vert \bm{a} - \bm{b} \Vert_1 \leq \beta & \\
\beta \cdot \left( \Vert \bm{a} - \bm{b} \Vert_1 - \frac{\beta}{2} \right) \quad \text{otherwise} &
\end{cases}
\label{eq:huber_loss}
\end{equation}

As discussed in Section~\ref{sec:related_work}, we decouple the min-max formulation of our objective given by Eq.~\ref{eq:our_learning_objective} and iterate between: (1) generating adversarial image $\hat{\bm{x}}$; and (2) minimizing the logit loss between $f_\theta\left( g_\phi(\hat{\bm{x}}) \right)$ and $f_\theta(\bm{x})$.
To generate adversarial images, we use $\ell_\infty$-PGD to maximize the cross entropy loss of the composite model $f_\theta(g_\phi(\bm{x}))$.
Note that while learning $\phi$, we backpropagate gradients through the frozen classifier $f_\theta$ without updating $\theta$.
Furthermore, our algorithm is agnostic to the selection of the pre-trained classifier, as further discussed in Section~\ref{sec:experiments}.

The network architecture of our RTFE is a modified UNet~\cite{ronneberger2015u} based on the proposal of~\cite{liao2018defense}.
We replace bilinear interpolation with nearest neighbor resize convolutions as this is known to reduce checkerboard artifacts~\cite{odena2016deconvolution}.
Additionally, we replace the final layer with a pixel convolution followed by a sigmoid layer.
Without this modification, the network frequently produced outputs with invalid pixel values.
We also tested using the denoising autoencoder architecture from~\cite{liao2018defense}, but found it's performance on clean images was worse than that of the UNet while offering similar robustness.

One notable property is that RTFEs trained on a single model are not likely to distinguish between examples exploiting \quotes{bugs} from those exploiting \quotes{features}, as both contribute to the guide network's vulnerability. Thus, the transferability of these RTFEs will be damaged as it is learning to protect against \quotes{bug} vulnerabilities that are not shared by other models. However, if the RTFE was instead trained on an ensemble of frozen guide models, we hypothesize it will better differentiate between \quotes{feature} and \quotes{bug} vulnerabilities, as the \quotes{features} are shared between the models it learns from, while the \quotes{bugs} are not. Thus, we invert the process proposed by \cite{nakkiran2019discussion}, which creates adversarial examples that exclusively exploit bugs, to generate adversarial examples that are more likely to exploit features.

\subsection{Robust Feature Extraction with Ensembles}
\label{sec:ensembling}

\begin{figure}
\centering
\includegraphics[width=0.75\columnwidth]{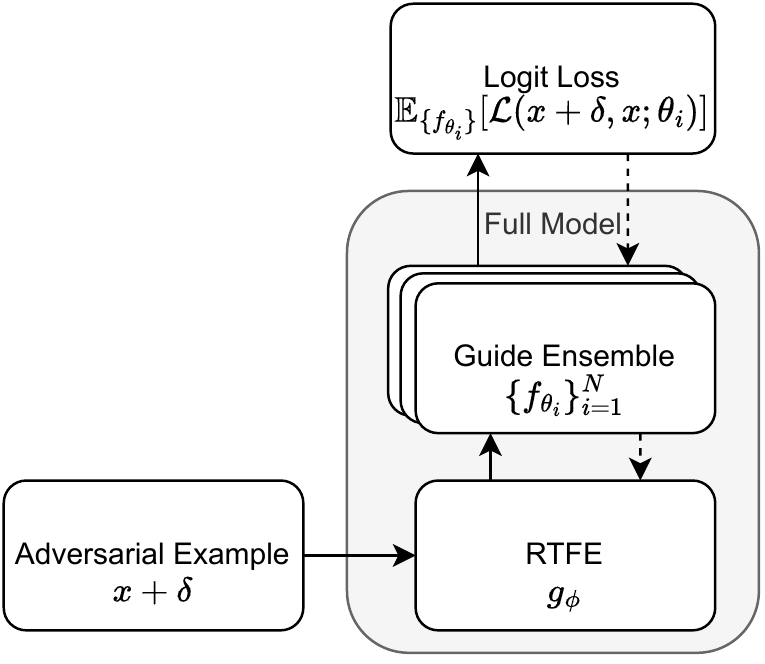}
\caption{We replace the single guide model $f_\theta$ with an ensemble of guide models $\{f_{\theta_i} \}^N_{i=1}$ to train an RTFE using the objective in Eq.~\ref{eq:ensemble_objective}.
Gradients are backpropagated through all frozen classifiers, but only used to update $\phi$. Adversarial examples are generated using PGD and Eq.~\ref{eq:ensemble_formulation}.}
\label{fig:guide_ensemble}
\end{figure}

In~\cite{nakkiran2019discussion}, the authors generate non-transferable adversarial examples using the attack objective given by Eq.~\ref{eq:bug_formulation}.
For an ensemble of $N$ independently trained models each parameterized by $\theta_i$,
this objective finds a perturbation that:
(1) minimizes the cross entropy loss with respect to an incorrect target class $\bm{y}_\text{targ}$ for model $f_{\vartheta}$; and (2) minimizes the expected standard loss with respect to the correct class $\bm{y}$ over an ensemble of models $\{f_{\theta_i}\}^N_{i=1}$ not containing $f_{\vartheta}$.
This has the effect of ignoring feature vulnerabilities while exploiting the bug vulnerabilities of model $f_{\vartheta}$, as bug vulnerabilities decrease the targeted adversarial loss with respect to $\bm{y}_\text{targ}$ for model $f_{\vartheta}$ without impacting the standard loss with respect to $\bm{y}$ for ensemble $\{f_{\theta_i}\}^N_{i=1}$.
To validate the assumption that the adversarial examples generated from this procedure do not exploit \quotes{feature} vulnerabilities,~\cite{nakkiran2019discussion} demonstrate that their adversarial examples do not transfer to unique classifiers trained outside the ensemble.
\begin{equation}
\Loss{\bm{x}+ \delta}{\bm{y}_\text{targ}}{\vartheta}{CE} + \mathbb{E}_{\{f_{\theta_i}\}^N_{i=1}} [\Loss{\bm{x} + \delta}{\groundTruth}{\theta_i}{CE}]
\label{eq:bug_formulation}
\end{equation}

To increase the transferability of an RTFE, we build from this procedure by reversing the logic to generate adversarial examples that are more likely to exploit \quotes{feature} vulnerabilities while avoiding \quotes{bug} vulnerabilities.
To do so, we use Eq.~\ref{eq:ensemble_formulation} to \textit{maximize} the standard loss with respect to $\bm{y}$ over ensemble $\{f_{\theta_i}\}^N_{i=1}$.
This has the effect of ignoring bugs, as the generated adversarial examples, by construction, induce erroneous predictions for all models in ensemble $\{f_{\theta_i}\}^N_{i=1}$, and are thus not unique to any individual model.
\begin{equation}
    \max_\delta \mathbb{E}_{\{f_{\theta_i}\}^N_{i=1}}  [\Loss{\bm{x}+\delta}{\bm{y}}{\theta_i}{CE})]
    \label{eq:ensemble_formulation}
\end{equation}

Recall that we use the composition of RTFE $g_\phi$ and guide model $f_\theta$ to generate adversarial examples during training with the process depicted in Figure~\ref{fig:train_rtfe}.
To learn an RTFE using an ensemble of guide models, we replace guide model $f_\theta$ with an ensemble of independently pre-trained classifiers $\{f_{\theta_i}\}^N_{i=1}$, as depicted in Figure~\ref{fig:guide_ensemble}.
Thus, for RTFE $g_\phi$ we solve the optimization objective given by Eq.~\ref{eq:ensemble_objective}, where adversarial example  $\hat{\bm{x}}$ is approximated using Eq.~\ref{eq:ensemble_formulation} from the composition of RTFE $g_\phi$ and ensemble $\{f_{\theta_i}\}^N_{i=1}$.
\begin{equation}
\min_\phi \mathbb{E}_{\{f_{\theta_i}\}^N_{i=1}}  [\Loss{\hat{\bm{x}}}{\bm{y}}{\phi,\theta_i}{logit}]
\label{eq:ensemble_objective}
\end{equation}

\section{Experiments and Results}
\label{sec:experiments}

\newcommand{\stepsize}{$\frac{1.25\epsilon}{255}$}
\newcommand{\steps}{20 steps }
\newcommand{\restarts}{5 restarts }

We empirically evaluate the performance of our algorithm in three scenarios: (1) the standard white box threat model; (2) a white box cross-model transfer scenario; and (3) a white box cross-dataset transfer scenario.
We compare our approach against adversarial training (AT)~\cite{madry2017towards} and joint adversarial training (JAT)~\cite{zhou2021improving}, and omit comparisons to the approach of \cite{liao2018defense} as it is shown to be ineffective against a white box adversary \cite{Athalye2018OnTR}.

Our models are trained to classify images using the CIFAR10 and CIFAR100 datasets~\cite{krizhevsky2009learning}, the most popular datasets for robustness research \cite{croce2021robustbench}. When training our RTFEs, we maintain the hyperparameters reported in~\cite{madry2017towards}, aside from a shortened training time of 60 epochs as we found that training longer did not improve the robustness of our models.
For an even comparison, we use a UNet autoencoder as the architecture for all defenses that require a pre-processing network.
We use ResNet18 and MobileNetV2 as our pre-trained classifiers and evaluate white box robustness using the projected gradient descent (PGD) attack with a step size of \stepsize ~for \steps ~\cite{madry2017towards}.
Furthermore, we evaluate at multiple values of $\epsilon$ such that the adversarial perturbation $\delta$ of the adversary is bounded by $\Vert \delta \Vert_\infty \leq \frac{\epsilon}{255}$, where $\epsilon \in \{0, 2, 4, 8 \} $. For convenience, we use $\epsilon = 0$ to denote the performance on unattacked images. We also include evaluations with stronger attacks in Appendix~\ref{appendix:gradient_obfuscation} to confirm that our defense does not suffer from gradient obfuscation.
All evaluations are performed on an AMD MI100 accelerator using PyTorch~\cite{adam2019pytorch}.
Additionally, we follow the best practices outlined in~\cite{carlini2019evaluating} by: (1) backpropagating through any pre-processing mechanisms applied to the inputs of a given classifier, including the defense; and (2) reporting all successfully attacked images at some $\epsilon$ as being successfully attacked for all larger $\epsilon$.

\begin{figure}[t]
    \centering
    \includegraphics[width=0.9\columnwidth]{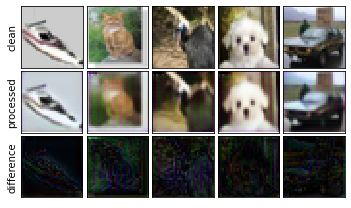}
    \caption{We provide a random sample of clean input images $\bm{x}$, the same images processed by an RTFE $g_\phi(\bm{x})$, and the absolute pixel difference between the two.
    }
    \label{fig:rtfe_output}
\end{figure}

\subsection{White Box Pre-Processing Defense}
\label{sec:white_box_evaluation}

\begin{figure}[t]
\centering
\subfloat[CIFAR10]{\includegraphics[width=0.5\linewidth]{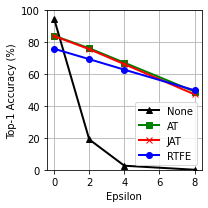}}
\subfloat[CIFAR100]{\includegraphics[width=0.5\linewidth]{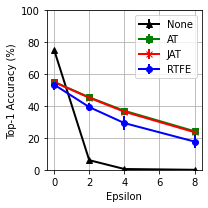}}
\caption{We visualize the performance trends on both the CIFAR10 and CIFAR100 datasets using a white box threat model at various $\epsilon$.}
\label{fig:standard_white_box}
\end{figure}

First, we evaluate our algorithm as a pre-processing defense using the standard white box threat model.
We train an RTFE using a frozen ResNet18 as our guide model. Recall that, in our algorithm, guide models are independently pre-trained on the target dataset under the standard learning objective.
During evaluation, we consider the adversarially trained RTFE and the pre-trained ResNet18 as a composite model.
For AT~\cite{madry2017towards} and JAT~\cite{zhou2021improving}, we maintain the hyperparameter configurations reported in the original papers to the extent possible.
We re-implement both algorithms and adversarially train both networks from scratch, which further ensures a fair comparison as different attack implementations and network architectures across frameworks could lead to inconsistencies.
As an additional baseline, we independently evaluate the pre-trained and undefended ResNet18 that was used as the guide model to train our RTFE.
We repeat each experiment 5 times using different random seeds and report the average top-1 test accuracy for both CIFAR10 and CIFAR100 in Table~\ref{tbl:std_white_box}.
In Figure~\ref{fig:standard_white_box}, we visualize the model performance trends and provide standard deviation error bars as we increase the attack perturbation bound $\epsilon$.
In both Table~\ref{tbl:std_white_box} and Figure~\ref{fig:standard_white_box}, we refer to the undefended ResNet18 as \quotes{None}.

We observe that, even without updating the undefended guide model, our RTFE provides a defense against an adaptive white box adversary that is comparable to both AT and JAT.
However, we observe that both JAT and AT have higher top-1 classification accuracy on unattacked images (\textit{i.e.}, $\epsilon=0$) on both datasets---a gap that is noticeably smaller on CIFAR100 than it is on CIFAR10.
This is expected as both JAT and AT train all layers of the model to be robust against adversarial examples.
In contrast, the model used to guide our RTFE is frozen during training such that only the RTFE is trained to be robust.
Surprisingly, our method had the highest robust accuracy on CIFAR10 when $\epsilon=8$.
We attribute this increase over AT to the additional model capacity resulting from prepending the RTFE to the ResNet18.

\begin{table}[t]
\centering
{\small
\begin{tabular}{cl|c|c|c|c|}
\cline{3-6}
                                         &   & $\epsilon=0$ & $\epsilon=2$ & $\epsilon=4$ & $\epsilon=8$ \\ \hline
\multicolumn{1}{|c|}{\multirow{4}{*}{CIFAR10}} & {None} &   94.8\% & 19.1\% & 2.6\% & 0.0\%  \\ \cline{2-6} 
\multicolumn{1}{|c|}{}                   & {AT} & 84.1\% & 76.3\% & 67.2\% & 49.1\% \\ \cline{2-6} 
\multicolumn{1}{|c|}{}                   & {JAT} & 83.9\%  & 75.7\% & 66.3\% & 47.3\% \\ \cline{2-6} 
\multicolumn{1}{|c|}{}                   & {RTFE} & 75.9\% & 69.4\% & 62.8\% & 49.9\% \\ \hline \hline
\multicolumn{1}{|c|}{\multirow{4}{*}{CIFAR100}} & {None} &   75.3\% & 6.0\% & 0.5\% & 0.0\%  \\ \cline{2-6} 
\multicolumn{1}{|c|}{}                   & {AT} & 55.2\% & 45.5\% & 37.0\% & 24.2\% \\ \cline{2-6} 
\multicolumn{1}{|c|}{}                   & {JAT} & 55.0\%  & 45.2\% & 36.4\% & 23.6\% \\ \cline{2-6} 
\multicolumn{1}{|c|}{}                   & {RTFE} & 53.4\% & 39.4\% & 29.3\% & 17.7\% \\ \hline
\end{tabular}}
\caption{We compare the performance of each defense against a white box adversary on both CIFAR10 and CIFAR100. We report the average top-1 classification accuracy over 5 independent runs.}
\label{tbl:std_white_box}
\end{table}

\subsection{Cross-Model Transfer Defense}
\label{sec:cross_model_exp}

\begin{figure}[t]
\centering
\subfloat[ResNet18]{\includegraphics[width=0.5\linewidth]{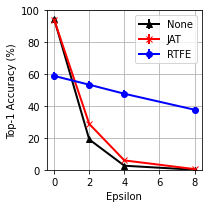}}
\subfloat[MobileNetV2]{\includegraphics[width=0.5\linewidth]{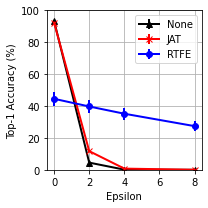}}
\caption{We visualize performance trends when both JAT and RTFE are trained on CIFAR10 using a ResNet18 as a guide model, then transfered cross-model to defend an undefended model independently trained on CIFAR10.}
\label{fig:cross_model_transfer}
\end{figure}

Next, we evaluate our algorithm as a cross-model transfer defense under the standard white box threat model.
To do so, we first train an RTFE using as our guide model a ResNet18 pre-trained on CIFAR10.
During evaluation, we use the resulting RTFE to defend two new models independently trained on the same dataset: (1) a new ResNet18 model trained from a different random initialization; and (2) a MobileNetV2 model.
We compare the performance of our algorithm against JAT, where the pre-processing model was jointly trained on CIFAR10 from scratch with a randomly initialized ResNet18 model.
In this cross-model transfer defense experiment, we are unable to compare against AT as there is no transferable defense mechanism.
We repeat each experiment 5 times using different random seeds and report the average top-1 classification accuracy in Table~\ref{tbl:cross_model_results}.
Furthermore, we visualize the trends as we increase $\epsilon$ in Figure~\ref{fig:cross_model_transfer}.
We observe that our RTFE significantly outperforms JAT when used as a transfer defense for both classifiers for all $\epsilon > 0$; however, we perform worse on unattacked images (\textit{i.e.}, $\epsilon=0$).
We discuss how we use ensembling to improve performance on unattacked images in Section~\ref{sec:optimizations_exp}.

\begin{table}[t]
\centering
{\small
\begin{tabular}{cl|c|c|c|c|}
\cline{3-6}
  &   & $\epsilon=0$ & $\epsilon=2$ & $\epsilon=4$ & $\epsilon=8$ \\ \hline
\multicolumn{1}{|c|}{\multirow{2}{*}{ResNet18}}  & {JAT} &   94.2\% & 28.8\% & 5.9\% & 0.4\%  \\ \cline{2-6}
\multicolumn{1}{|c|}{}                   & {RTFE} & 58.9\% & 53.4\% & 47.7\% & 37.7\% \\ \hline \hline
\multicolumn{1}{|c|}{\multirow{2}{*}{MobileNetV2}}  & {JAT} &   92.3\% & 11.7\% & 0.7\% & 0.0\%  \\ \cline{2-6}
\multicolumn{1}{|c|}{}                   & {RTFE} & 44.5\% & 39.7\% & 35.2\% & 27.4\% \\ \hline
\end{tabular}}
\caption{We compare the performance of our RTFE algorithm against JAT when transfered cross-model on CIFAR10. We use a ResNet18 as the guide model for both JAT and RTFE, then evaluate on an independently trained ResNet18 (top) and MobileNetV2 (bottom). We report the average top-1 test accuracy over 5 independent runs.}
\label{tbl:cross_model_results}
\end{table}

\subsection{Cross-Dataset Transfer Defense}
\label{sec:cross_dataset_exp}

\begin{figure}[t]
    \centering
	\subfloat[ResNet18]{\includegraphics[width=0.5\linewidth]{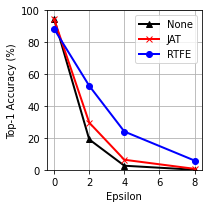}}
	\subfloat[MobileNetV2]{\includegraphics[width=0.5\linewidth]{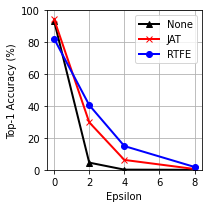}}
    \caption{We visualize performance trends when each algorithm is first trained on CIFAR100, then used as a cross-dataset transfer defense on CIFAR10 at various $\epsilon$.}
    \label{fig:cross_dataset_transferability}
\end{figure}

Here, we evaluate our algorithm as a cross-dataset transfer defense using the standard white box threat model.
To do so, we first train RTFEs using as our guide models a ResNet18 and MobileNetV2 pre-trained on CIFAR100.
During evaluation, we use the resulting RTFE to defend a model of the same architecture separately trained on CIFAR10.
We again compare the performance of our RTFE algorithm against JAT, where the pre-processing model was jointly trained from scratch on CIFAR100 with a randomly initialized ResNet18 or MobileNetV2, respectively.  
We repeat each cross-dataset transfer defense experiment 5 times using different random seeds and report the maximum observed top-1 classification accuracy in Table~\ref{tbl:cross_dataset_results}.
Furthermore, we visualize the trends in Figure~\ref{fig:cross_dataset_transferability} and provide standard deviation errors bars as we increase $\epsilon$.
As before, we see that our RTFE outperforms JAT on both classifiers for all $\epsilon > 0$, yet performs worse on unattacked images  (\textit{i.e.}, $\epsilon=0$).
Surprisingly, we also observe that this cross-dataset transfer defense significantly outperforms cross-model with lower values of $\epsilon$.
We conjecture this can be attributed to the transferability of features shared between datasets and is in part due to the increased complexity of the source dataset when compared to the target dataset.
We direct the interested reader to~\cite{salman2020adversarially}, which explores cross-dataset transferability within the scope of transfer learning.

\begin{table}[t]
\centering
{\small
\begin{tabular}{cl|c|c|c|c|}
\cline{3-6}
                                         &   & $\epsilon=0$ & $\epsilon=2$ & $\epsilon=4$ & $\epsilon=8$ \\ \hline
\multicolumn{1}{|c|}{\multirow{2}{*}{ResNet18}}  & {JAT} &   94.3\% & 27.0\% & 5.2\% & 0.4\%  \\ \cline{2-6}
\multicolumn{1}{|c|}{}                   & {RTFE} & 88.4\% & 52.7\% & 24.0\% & 5.8\% \\ \hline \hline
\multicolumn{1}{|c|}{\multirow{2}{*}{MobileNetV2}} & {JAT} & 94.5\% & 29.8\% & 6.2\% & 0.5\% \\ \cline{2-6} 
\multicolumn{1}{|c|}{}                   & {RTFE} & 82.2\%  & 40.6\% & 14.9\% & 1.8\% \\ \hline
\end{tabular}}
\caption{We compare the performance of our RTFE against JAT when transfered cross-dataset from CIFAR100 to CIFAR10. We use ResNet18 (top) and MobileNetV2 (bottom) models trained on CIFAR100 as guides, before transfering to the same architectures trained on CIFAR10. We report the best top-1 test accuracy over 5 independent runs.}
\label{tbl:cross_dataset_results}
\end{table}

\subsection{Impact of Model Size on Transferability}
\label{sec:model_size_exp}

It is clear that network architecture has a significant impact on the performance of our RTFE when used as a cross-model defense. In Table~\ref{tab:guide_size_investigation}, we report the top-1 test accuracy observed when transferring an RTFE trained and transfered to ResNet models of varied capacity. We repeat each experiment 3 times using different random seeds and report the average.
When further investigating the impact of the guide model architecture on transferability, we observe that transfer defenses, unlike transfer attacks \cite{lei2020towards}, do not perform worse when moving from a larger model to a smaller one. We conjecture that this is due to the impact of \quotes{bug} vulnerabilties, which asymetrically effect both transfer attacks and  defenses. Whereas an RTFE may be affected by \quotes{bugs} in both the guide and defended models, a transfered adversarial example is only able to interact with \quotes{bugs} in the model used to generate it.

\begin{table}[t]
\centering
\begin{tabular}{|l|l|c|c|}
\hline
Guide Model & Defended Model & $\epsilon=0$ & $\epsilon=8$ \\ \hline \hline
 ResNet50 & ResNet34 & 64.9\% & 39.5\% \\ \hline
 ResNet50 & ResNet18 & 59.5\% & 38.5\% \\ \hline \hline
 ResNet18 & ResNet50 & 46.3\% & 28.0\% \\ \hline
 ResNet18 & ResNet34 & 61.8\% & 39.5\% \\ \hline
\end{tabular}
\caption{We evaluate the performance of our cross-model transfer defense on CIFAR10 as we vary model capacity.}
\label{tab:guide_size_investigation}
\end{table}

\subsection{Improving Transferability with Ensembles}
\label{sec:optimizations_exp}

\begin{figure}[b!]
\centering
\subfloat[$\epsilon=0$]{\includegraphics[width=0.5\linewidth]{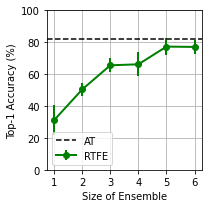}}
\subfloat[$\epsilon=8$]{\includegraphics[width=0.5\linewidth]{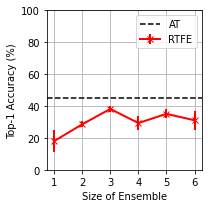}}
\caption{We visualize the performance trends of our RTFE on clean images (left) and attacked images (right) when used as a cross-model transfer defense against a white box threat model on CIFAR10 as we increase the size of the ensemble.}
\label{fig:ensemble_size}
\end{figure}

To examine our hypothesis that training our RTFE with an ensemble of guide models will improve transferability, we return to the cross-model transfer defense.
Using our algorithm detailed in Section~\ref{sec:ensembling}, we train our RTFE using an ensemble created by randomly selecting $m$ models without replacement from a pool of $n$ independently pre-trained ResNet18, MobileNetV2, and SqueezeNet \cite{iandola2016squeezenet} classifiers. In our experiments, our pool consists of 2 ResNet18 models, 2 MobileNetV2 models, and 2 SqueezeNet models such that $n=6$.
During evaluation, we use the resulting RTFE to defend an independently trained VGG16 \cite{simonyan2015very}.
Note that this architecture is not included in the pool of ensembles to control for the impact of model similarity on transferability.
Using a white box threat model, we evaluate the performance of our RTFE trained using increased ensemble sizes such that $m \in \{1, 2, 3, 4, 5, 6\}$.
We repeat each experiment 5 times and visualize the trends in Figure~\ref{fig:ensemble_size}, where we plot the mean and standard deviation as we increase $\epsilon$.
We also plot the performance of an adversarially trained VGG16 model as a baseline.

When transferred to defend a model with a previously unseen network architecture, we observe that RTFEs trained using a single guide model (\textit{i.e.}, $m=1$) were noticeably less performant than those trained with 2 or more guide models (\textit{i.e.}, $m \geq 2$).
Furthermore, we observe that, while the transfer performance on unattacked images ($\epsilon=0$) linearly increases with the size of the ensemble, the transfer performance on attacked images ($\epsilon=8$) saturates around an ensemble of 3.
We conjecture this drastic increase in the performance of our RTFE on clean images can be attributed in part to being trained using more useful directions for learning (\textit{i.e.}, adversarial examples primarily exploiting feature vulnerabilities).
However, bug vulnerabilities can still be exploited by malicious white box adversaries during inference.
Thus, we do not see as strong a correlation between ensemble size and robust accuracy as with ensemble size and clean accuracy.

\section{Conclusion and Future Work}
\label{sec:conclusion}

In this work, we explored the use of transferable defense mechanisms against adaptive white box adversaries.
We hypothesized that a deep neural network can take as input an adversarial example and give as output a robust representation that can be passed to independently pre-trained classifiers that are otherwise ineffective against a white box adverary.
Our results show that our robust transferable feature extractor (RTFE) increases the robustness of otherwise undefended classifiers in all scenarios investigated, thus validating this hypothesis.
We summarize the following conclusions:
\begin{itemize}
\item An RTFE trained using our algorithm introduced in Section 3.2 can be used as a pre-processing defense against a white box adversary to provide robustness comparable to adversarial training when used by its guide model.
\item Once trained, an RTFE can be successfully used as a cross-model transfer defense, where it provides adversarial robustness to another model independently pre-trained on the same dataset.
\item An RTFE can be successfully used as a cross-dataset transfer defense, where it provides adversarial robustness to another model independently pre-trained on a different dataset of similar or lesser complexity.
\item Training an RTFE using an ensemble of guide models increases its effectiveness as a cross-model transfer defense, where the performance on unattacked images increases linearly with ensemble size and the performance on attacked images saturates early.
\end{itemize}
Thus, once trained, RTFEs can be reused to defend models that have not yet been created and can amortize the costs of adversarial training over multiple non-robust models that are otherwise undefended.

Several interesting directions remain open for exploration. In our experiments, we find that training to prioritize \textit{feature} adversarial examples significantly outperforms training on a single classifier. It remains to be seen whether one can effectively utilize \textit{feature} adversarial examples to improve regular adversarial training. Furthermore, our experiments show that RTFEs can be successfully reused across datasets for the same task. It may follow that one could reuse an RTFE across different tasks (\textit{e.g.}, image classification to object detection). We leave these investigations for future work.

\section*{Acknowledgements}
We would like to thank the AMD Software Technology and Architecture team for their insightful discussions and infrastructure support. This work was supported in part by Mitacs through the Mitacs Accelerate program. 
\newline

\noindent © 2022 Advanced Micro Devices, Inc.  All rights reserved.
AMD, the AMD Arrow logo, Radeon, and combinations thereof are trademarks of Advanced Micro Devices, Inc. Other product names used in this publication are for identification purposes only and may be trademarks of their respective companies.

\bibliography{citations.bib}

\begin{thebibliography}{23}
\providecommand{\natexlab}[1]{#1}

\bibitem[{Athalye and Carlini(2018)}]{Athalye2018OnTR}
Athalye, A.; and Carlini, N. 2018.
\newblock On the Robustness of the CVPR 2018 White-Box Adversarial Example
  Defenses.
\newblock arXiv:1804.03286.

\bibitem[{Athalye, Carlini, and Wagner(2018)}]{athalye2018obfuscated}
Athalye, A.; Carlini, N.; and Wagner, D. 2018.
\newblock Obfuscated Gradients Give a False Sense of Security: Circumventing
  Defenses to Adversarial Examples.
\newblock In \emph{Proceedings of the 35th International Conference on Machine
  Learning, ICML 2018}.

\bibitem[{Carlini et~al.(2019)Carlini, Athalye, Papernot, Brendel, Rauber,
  Tsipras, Goodfellow, Madry, and Kurakin}]{carlini2019evaluating}
Carlini, N.; Athalye, A.; Papernot, N.; Brendel, W.; Rauber, J.; Tsipras, D.;
  Goodfellow, I.; Madry, A.; and Kurakin, A. 2019.
\newblock On evaluating adversarial robustness.
\newblock arXiv:1902.06705.

\bibitem[{Croce et~al.(2021)Croce, Andriushchenko, Sehwag, Debenedetti,
  Flammarion, Chiang, Mittal, and Hein}]{croce2021robustbench}
Croce, F.; Andriushchenko, M.; Sehwag, V.; Debenedetti, E.; Flammarion, N.;
  Chiang, M.; Mittal, P.; and Hein, M. 2021.
\newblock RobustBench: a standardized adversarial robustness benchmark.
\newblock In \emph{Thirty-fifth Conference on Neural Information Processing
  Systems Datasets and Benchmarks Track (Round 2)}.

\bibitem[{Guo et~al.(2018)Guo, Rana, Cisse, and van~der
  Maaten}]{guo2018countering}
Guo, C.; Rana, M.; Cisse, M.; and van~der Maaten, L. 2018.
\newblock Countering Adversarial Images using Input Transformations.
\newblock In \emph{International Conference on Learning Representations}.

\bibitem[{Huber(1992)}]{huber1992robust}
Huber, P.~J. 1992.
\newblock Robust estimation of a location parameter.
\newblock In \emph{Breakthroughs in statistics}, 492--518. Springer.

\bibitem[{Iandola et~al.(2016)Iandola, Moskewicz, Ashraf, Han, Dally, and
  Keutzer}]{iandola2016squeezenet}
Iandola, F.~N.; Moskewicz, M.~W.; Ashraf, K.; Han, S.; Dally, W.~J.; and
  Keutzer, K. 2016.
\newblock SqueezeNet: AlexNet-level accuracy with 50x fewer parameters and
  {\textless}1MB model size.
\newblock arXiv:1602.07360.

\bibitem[{Ilyas et~al.(2019)Ilyas, Santurkar, Tsipras, Engstrom, Tran, and
  Madry}]{ilyas2019adversarial}
Ilyas, A.; Santurkar, S.; Tsipras, D.; Engstrom, L.; Tran, B.; and Madry, A.
  2019.
\newblock Adversarial Examples Are Not Bugs, They Are Features.
\newblock In Wallach, H.; Larochelle, H.; Beygelzimer, A.; d\textquotesingle
  Alch\'{e}-Buc, F.; Fox, E.; and Garnett, R., eds., \emph{Advances in Neural
  Information Processing Systems}, volume~32. Curran Associates, Inc.

\bibitem[{Joshi et~al.(2022)Joshi, Kataria, Shao, Zelasko, Villalba, Khudanpur,
  and Dehak}]{joshi2022defense}
Joshi, S.; Kataria, S.; Shao, Y.; Zelasko, P.; Villalba, J.; Khudanpur, S.; and
  Dehak, N. 2022.
\newblock Defense against Adversarial Attacks on Hybrid Speech Recognition
  using Joint Adversarial Fine-tuning with Denoiser.
\newblock arXiv:2204.03851.

\bibitem[{Krizhevsky, Hinton et~al.(2009)}]{krizhevsky2009learning}
Krizhevsky, A.; Hinton, G.; et~al. 2009.
\newblock Learning multiple layers of features from tiny images.
\newblock Technical report, University of Toronto.

\bibitem[{Liao et~al.(2018)Liao, Liang, Dong, Pang, Zhu, and
  Hu}]{liao2018defense}
Liao, F.; Liang, M.; Dong, Y.; Pang, T.; Zhu, J.; and Hu, X. 2018.
\newblock Defense Against Adversarial Attacks Using High-Level Representation
  Guided Denoiser.
\newblock \emph{2018 IEEE/CVF Conference on Computer Vision and Pattern
  Recognition}, 1778--1787.

\bibitem[{Madry et~al.(2018)Madry, Makelov, Schmidt, Tsipras, and
  Vladu}]{madry2017towards}
Madry, A.; Makelov, A.; Schmidt, L.; Tsipras, D.; and Vladu, A. 2018.
\newblock Towards Deep Learning Models Resistant to Adversarial Attacks.
\newblock In \emph{International Conference on Learning Representations}.

\bibitem[{Nakkiran(2019)}]{nakkiran2019discussion}
Nakkiran, P. 2019.
\newblock A Discussion of 'Adversarial Examples Are Not Bugs, They Are
  Features': Adversarial Examples are Just Bugs, Too.
\newblock \emph{Distill}.
\newblock Https://distill.pub/2019/advex-bugs-discussion/response-5.

\bibitem[{Odena, Dumoulin, and Olah(2016)}]{odena2016deconvolution}
Odena, A.; Dumoulin, V.; and Olah, C. 2016.
\newblock Deconvolution and checkerboard artifacts.
\newblock \emph{Distill}, 1(10): e3.

\bibitem[{Paszke et~al.(2019)Paszke, Gross, Massa, Lerer, Bradbury, Chanan,
  Killeen, Lin, Gimelshein, Antiga, Desmaison, Kopf, Yang, DeVito, Raison,
  Tejani, Chilamkurthy, Steiner, Fang, Bai, and Chintala}]{adam2019pytorch}
Paszke, A.; Gross, S.; Massa, F.; Lerer, A.; Bradbury, J.; Chanan, G.; Killeen,
  T.; Lin, Z.; Gimelshein, N.; Antiga, L.; Desmaison, A.; Kopf, A.; Yang, E.;
  DeVito, Z.; Raison, M.; Tejani, A.; Chilamkurthy, S.; Steiner, B.; Fang, L.;
  Bai, J.; and Chintala, S. 2019.
\newblock PyTorch: An Imperative Style, High-Performance Deep Learning Library.
\newblock In Wallach, H.; Larochelle, H.; Beygelzimer, A.; d\textquotesingle
  Alch\'{e}-Buc, F.; Fox, E.; and Garnett, R., eds., \emph{Advances in Neural
  Information Processing Systems 32}, 8024--8035. Curran Associates, Inc.

\bibitem[{Raghunathan et~al.(2019)Raghunathan, Xie, Yang, Duchi, and
  Liang}]{raghunathan2019adversarial}
Raghunathan, A.; Xie, S.~M.; Yang, F.; Duchi, J.~C.; and Liang, P. 2019.
\newblock Adversarial training can hurt generalization.
\newblock arXiv:1906.06032.

\bibitem[{Ronneberger, Fischer, and Brox(2015)}]{ronneberger2015u}
Ronneberger, O.; Fischer, P.; and Brox, T. 2015.
\newblock U-net: Convolutional networks for biomedical image segmentation.
\newblock In \emph{International Conference on Medical image computing and
  computer-assisted intervention}, 234--241. Springer.

\bibitem[{Salman et~al.(2020)Salman, Ilyas, Engstrom, Kapoor, and
  Madry}]{salman2020adversarially}
Salman, H.; Ilyas, A.; Engstrom, L.; Kapoor, A.; and Madry, A. 2020.
\newblock Do adversarially robust imagenet models transfer better?
\newblock \emph{Advances in Neural Information Processing Systems}, 33:
  3533--3545.

\bibitem[{Simonyan and Zisserman(2015)}]{simonyan2015very}
Simonyan, K.; and Zisserman, A. 2015.
\newblock Very Deep Convolutional Networks for Large-Scale Image Recognition.
\newblock In Bengio, Y.; and LeCun, Y., eds., \emph{3rd International
  Conference on Learning Representations, {ICLR} 2015, San Diego, CA, USA, May
  7-9, 2015, Conference Track Proceedings}.

\bibitem[{Su et~al.(2018)Su, Zhang, Chen, Yi, Chen, and Gao}]{su2018robustness}
Su, D.; Zhang, H.; Chen, H.; Yi, J.; Chen, P.-Y.; and Gao, Y. 2018.
\newblock Is Robustness the Cost of Accuracy?--A Comprehensive Study on the
  Robustness of 18 Deep Image Classification Models.
\newblock In \emph{Proceedings of the European Conference on Computer Vision
  (ECCV)}, 631--648.

\bibitem[{Tsipras et~al.(2019)Tsipras, Santurkar, Engstrom, Turner, and
  Madry}]{tsipras2018robustness}
Tsipras, D.; Santurkar, S.; Engstrom, L.; Turner, A.; and Madry, A. 2019.
\newblock Robustness May Be at Odds with Accuracy.
\newblock In \emph{International Conference on Learning Representations}.

\bibitem[{Wu and Zhu(2020)}]{lei2020towards}
Wu, L.; and Zhu, Z. 2020.
\newblock Towards Understanding and Improving the Transferability of
  Adversarial Examples in Deep Neural Networks.
\newblock In Pan, S.~J.; and Sugiyama, M., eds., \emph{Proceedings of The 12th
  Asian Conference on Machine Learning}, volume 129 of \emph{Proceedings of
  Machine Learning Research}, 837--850. PMLR.

\bibitem[{Zhou et~al.(2021)Zhou, Wang, Gao, Han, Yu, Wang, and
  Liu}]{zhou2021improving}
Zhou, D.; Wang, N.; Gao, X.; Han, B.; Yu, J.; Wang, X.; and Liu, T. 2021.
\newblock Improving white-box robustness of pre-processing defenses via joint
  adversarial training.
\newblock arXiv:2106.05453.

\end{thebibliography}

\vfill
\pagebreak

\appendix

\section{Gradient Obfuscation}
\label{appendix:gradient_obfuscation}

To confirm that our robust transferable feature extractors (RTFEs) do not suffer from gradient obfuscation, we follow the investigations of~\cite{carlini2019evaluating}.
We start by noting that the UNet architecture used in our experiments does not contain randomized or non-differentiable layers; thus, gradient obfuscation is unlikely~\cite{carlini2019evaluating}.
However, we confirm its absence by showing that: (1) single-step attacks perform worse than iterative attacks; and (2) incrementally increasing $\epsilon$ will inevitably cause model performance to fall below random.
In these experiments, we use a more exhaustive projected gradient descent (PGD) attack to confirm that our choice of hyperparameters were valid for the experiments detailed in Section~\ref{sec:experiments}.

We compare the performance of three configurations of PGD in the standard white box scenario on CIFAR10:
\begin{enumerate}
	\item A weak attack using a single step of PGD with $\alpha=1.25\epsilon$
	\item A strong attack that uses PGD with 100 steps, 5 restarts, and a step size of $\frac{1.25\epsilon}{100}$
	\item The attack we used in Section~\ref{sec:white_box_evaluation}, which uses 20 steps and stepsize $\frac{1.25\epsilon}{100}$.
\end{enumerate}
All RTFEs are trained and tested on ResNet18 classifiers with the same weights used during training and testing.
We evaluate at $\epsilon \in \{0,2,4,8,16,32\}$ to show that the top-1 accuracy converges to random chance given a large enough $\ell_\infty$ bound on the adversary.
Recall that we use $\epsilon=0$ to denote the performance on unattacked images.
We visualize the model performance trends and provide standard deviation bars as we increase the $\epsilon$.

\begin{figure}[ht]
\centering
    \includegraphics[width=0.8\columnwidth]{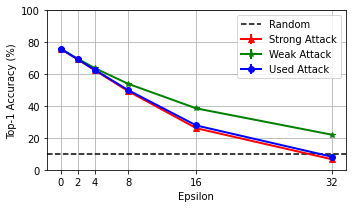}
    \caption{We confirm that our RTFE does not suffer from gradient obfuscation by incrementally increasing $\epsilon$. Using ResNet18 as our guide model for CIFAR10, we show that the top-1 accuracy converges to random chance with a large enough bound on the adversary.}
    \label{fig:gradient_obfuscation}
\end{figure}

From examining Figure~\ref{fig:gradient_obfuscation}, it is apparent that the single-step attack performs worse than both iterative approaches, and that at $\epsilon=32$ the model performs worse than a random classifier (\textit{i.e.}, top-1 accuracy less than $10\%$) on both iterative attacks.
Additionally, increasing the number of steps from 20 to 100, and adding additional repeats, only marginally increased the attack success at $\epsilon=8$ indicating that our choice of hyperparameters was sufficient.

\section{Evaluating RTFEs trained on Ensembles in the Standard White Box Threat Model}

We demonstrate that the increased transferability of an RTFE trained on an ensemble is not an artifact of training a generally more performant RTFE.
Consider two RTFEs $g$ and $g^\prime$ that provide $35\%$ and $30\%$ robustness, respectively, when transferred cross-model to defend against a bounded white box adversary.
If $g$ and $g^\prime$ also provide non-transfer robustness of $65\%$ and $40\%$, respectively, then it is unclear whether $g$ is learning a more transferable representation, as the performance drop is significantly more severe than $g^\prime$ when transferred.
Thus, the natural question arises: \textit{do RTFEs trained on ensembles perform similarly to those trained on a single model while still being more transferable?}
We investigate this question with the following additional experimentation.

\begin{figure}[ht]
\centering
    \subfloat[Non-Transfer]{\includegraphics[width=0.5\columnwidth]{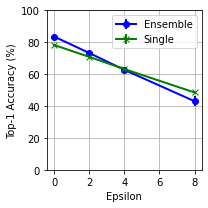}}
    \subfloat[Cross-Model]{\includegraphics[width=0.5\columnwidth]{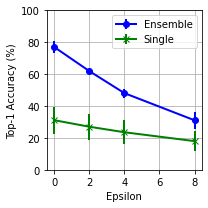}}
    \caption{We compare RTFEs trained using a single guide model against those trained using an ensemble of guide models. We evaluate the performance of these RTFEs when used as either a non-transfer defense or a cross-model transfer defense on CIFAR10. We observe that our ensemble training algorithm increases transferability.}
    \label{fig:ensemble_standard_performance}
\end{figure}

We evaluate a defended ensemble in the standard white box threat model using an RTFE trained on CIFAR10 with an ensemble of 6 guide models under the same conditions detailed in Section~\ref{sec:ensembling}.
We then have that RTFE defend the same ensemble used during training.
We compare the performance of RTFEs trained using a single model against RTFEs trained using an ensemble of models in both this non-transfer scenario as well as our cross-model transfer scenario.
To evaluate the robustness, we use the same attack hyperparameters detailed in Section~\ref{sec:white_box_evaluation} with the loss function detailed in Eq.~\ref{eq:ensemble_formulation}.
In Figure~\ref{fig:ensemble_standard_performance}, we visualize the performance trends in both the non-transfer and cross-model scenarios.
For the cross-model experiment, we evaluate using a frozen VGG16 guide model as discussed in Section~\ref{sec:optimizations_exp}.
We observe that, while RTFEs perform similarly on the model(s) used to guide training, there is a significant increase in robustness when used to guide novel models.
This provides further evidence that training on adversarial examples that exploit feature vulnerabilities further increases the transferability of RTFEs.

\end{document}